\documentclass[letterpaper, 10 pt, conference]{ieeeconf}  

\IEEEoverridecommandlockouts                              

\usepackage{graphicx}
\usepackage{algorithmic}
\usepackage[ruled,vlined]{algorithm2e}
\usepackage{multirow}
\usepackage{diagbox}
\usepackage{caption, subcaption, setspace}
\usepackage{amssymb}
\usepackage{float}

\usepackage{pifont}
\newcommand{\xmark}{\ding{55}}%

\title{\LARGE \bf
Towards End-to-End Deep Learning for Autonomous Racing: \\On Data Collection and a Unified Architecture \\for Steering and Throttle Prediction
}

\author{Shakti N. Wadekar$^{1}$, Benjamin J. Schwartz$^{1}$, Shyam S. Kannan$^{2}$, Manuel Mar$^{2}$, Rohan Kumar Manna$^{1}$,\\
 Vishnu Chellapandi$^{1}$, Daniel J. Gonzalez$^{3}$, Aly El Gamal$^{1}$ 
\thanks{$^{1}$School of Electrical and Computer Engineering, Purdue University, West Lafayette, IN 47907, USA. Email: {\tt\small (swadekar, schwartz, rmanna, cvp, elgamala)@purdue.edu}}%
\thanks{$^{2}$Computer Information Technology, Polytechnic Institute, Purdue University, West Lafayette, IN 47907, USA. Email: {\tt\small (kannan9, mmar)@purdue.edu}}%
\thanks{$^{3}$ Robotics Research Center, Department of Electrical Engineering and Computer Science, United States Military Academy, West Point, NY 10996, USA. Email: {\tt\small daniel.gonzalez@westpoint.edu.}}%
}

\begin{document}

\maketitle
\thispagestyle{empty}
\pagestyle{empty}

\begin{abstract}
Deep Neural Networks (DNNs) which are trained end-to-end have been successfully applied to solve complex problems that we have not been able to solve in past decades. Autonomous driving is one of the most complex problems which is yet to be completely solved and autonomous racing adds more complexity and exciting challenges to this problem. Towards the challenge of applying end-to-end learning to autonomous racing, this paper shows results on two aspects: (1) Analyzing the relationship between the driving data used for training and the maximum speed at which the DNN can be successfully applied for predicting steering angle, (2) Neural network architecture and training methodology for learning steering and throttle without any feedback or recurrent connections.

\end{abstract}

\section{Introduction}

High speed autonomous racing presents us with unique challenges that have gained recent attention after significant progress in urban autonomous driving. Investigating the deep-learning-based end-to-end learning solution to autonomous racing from a data-centric perspective is novel and no prior work has investigated this approach earlier to the best of our knowledge. Since this approach is new, we have to find and address the challenges associated with Data Collection, Deep Neural Network (DNN) architectures, as well as the suitable training methodology for this specific problem.

With Deep learning, it is typically the case that once we have a good dataset for a problem, multiple DNNs can achieve good performance. We call here the approach of exploring different architectures for solving the problem a model-centric approach. For example, an image recognition problem with the ILSVRC dataset \cite{ILSVRC15} can be solved with DNNs such as Alexnet \cite{krizhevsky2012imagenet}, ResNet \cite{he2016deep} and GoogleNet \cite{szegedy2015going}. In addition to that, these architectures can be successfully applied to similar other learning problems \cite{lin2014microsoft}. Since we can adopt the DNNs from other learning problems that are similar, to solve a new learning problem, a \textbf{data-centric approach to investigate the properties of a successful data collection strategy in the initial stages of tackling the considered new problem is necessary}. The urban autonomous driving problem is similar to autonomous racing with one major difference related to speed of operation; therefore, we can adopt DNNs from them and explore them more once we address the dataset challenges for autonomous racing, with particular emphasis on maximum speed encountered during both training and inference. 

First, with a data-centric approach, we explore the data with a fixed DNN similar to Nvidia's self-driving-car Neural Network Architecture \cite{bojarski2016end}. There are two reasons for selecting Nvidia's architecture, 1. End-to-end learning with this architecture was demonstrated to be successful, 2. Potential for achieving fast response time due to the simplified nature of the architecture, which is necessary for high-speed driving. \textbf{We here investigate successful data collection strategies for autonomous racing with a particular emphasis on the relationship between the amount of data collected and the maximum speed encountered during training for a steering prediction learning task}. Results are validated on two different tracks using the Unity simulation platform \cite{juliani2018unity}. On Track 1, we show results at constant speeds of 80 mph, 50 mph, and 30 mph, and on Track 2, we show results at constant speeds of 60 mph, 50 mph, and 30 mph. For speeds beyond 80 mph on Track 1 and 60 mph on Track 2, the throttle has to be varied requiring a throttle-dependent steering value. To investigate optimal data collection strategies in operating regimes with dependent steering and throttle values, we first need to solve the problem of joint steering and throttle prediction. Our second contribution solves this problem as outlined below. 

Second, \textbf{unlike previous literature, we show in this work that throttle prediction can be carried out via: 1. Learned convolutional layer weights from steering training without retraining them during throttle learning, and 2. Without using any feedback links \cite{Pan-RSS-18, 8460487} or recurrence (LSTM) \cite{8546189, 8099859, hecker2018end} in the deep neural network architecture}. This is achieved with a carefully designed training methodology explained in detail in Section V. We show that in order to jointly learn steering and throttle, we can use the same convolutional features learned during the steering training and only train subsequent fully connected layers dedicated for throttle inference. 
To validate the results, we demonstrate this on two different tracks. Video links given in section V results.

\section{Experimental Setup}

\subsection{Simulator}
A custom simulator was built by importing Udacity self-driving-car-sim\footnote{https://github.com/udacity/self-driving-car-sim} into the Unity platform \cite{juliani2018unity} and modifying it to have variable maximum speed as well as the ability to build new tracks. It uses the same car model which is provided by Udacity self-driving-car-sim. We built two new tracks (1) Track 1, which is similar to the Indy Moto Speedway (IMS) (2) Track 2, which is customly designed to have sharp right and left turns. 
Track 1 and Track 2 are $\approx1.6$ miles and $\approx2.3$ miles long, respectively.

\begin{figure}[ht]
\centering
\begin{minipage}[b]{0.45\linewidth}
\includegraphics[scale=0.16]{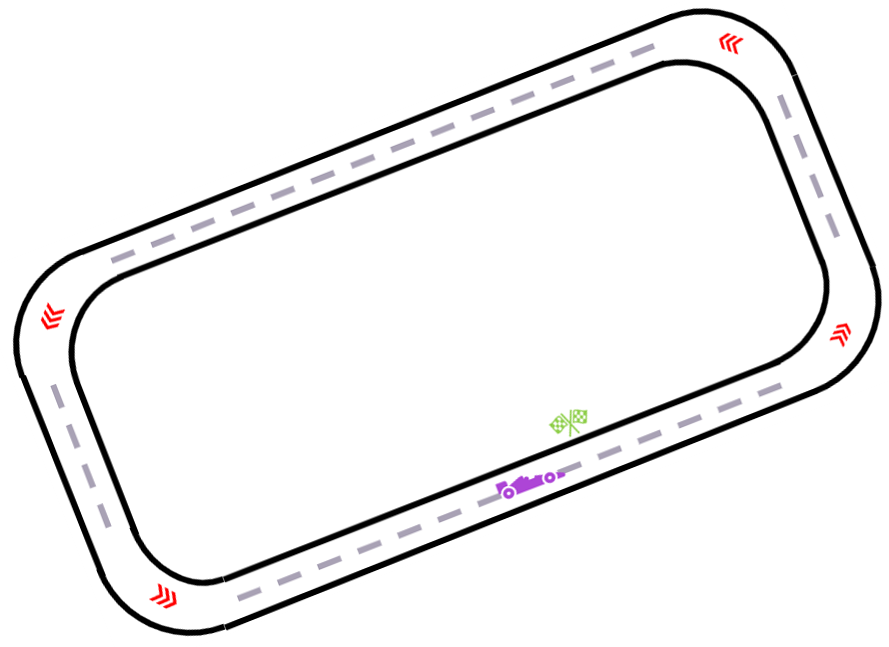}
\caption{Track 1: Akin IMS}
\label{fig:minipage1}
\end{minipage}
\quad
\begin{minipage}[b]{0.45\linewidth}
\includegraphics[scale=0.17]{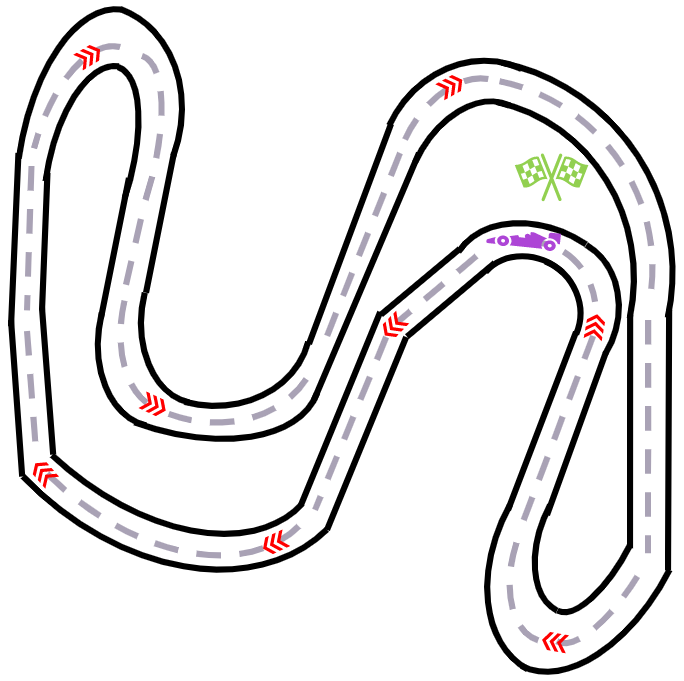}
\caption{Track 2: Custom track with sharp right and left turns}
\label{fig:minipage2}
\end{minipage}
\end{figure}

\begin{figure}[ht]
\centering
\begin{minipage}[b]{0.45\linewidth}
\includegraphics[scale=0.145]{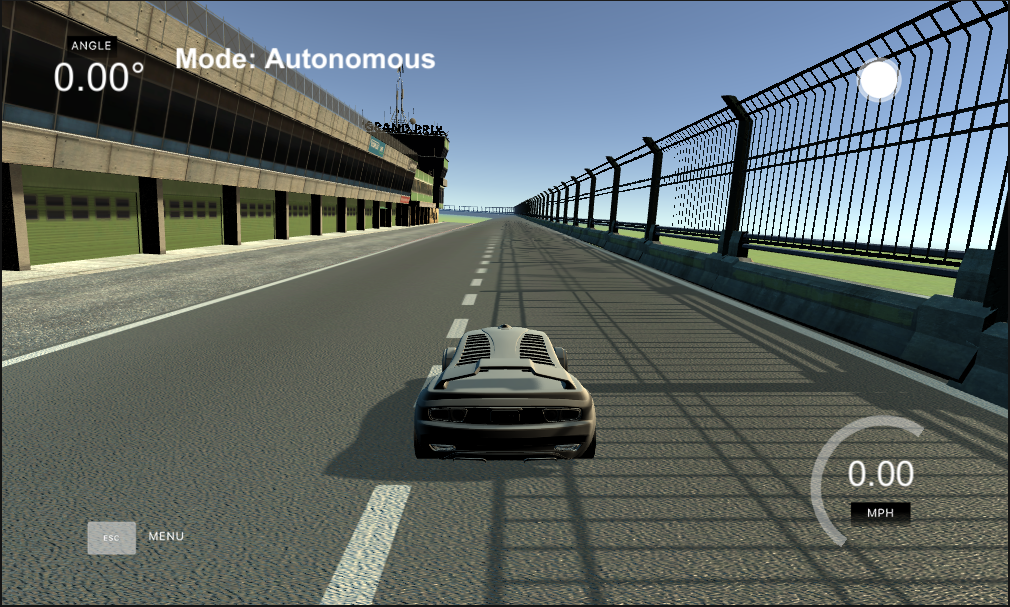}
\caption{Track 1: Simulator and car}
\label{fig:minipage3}
\end{minipage}
\quad
\begin{minipage}[b]{0.45\linewidth}
\includegraphics[scale=0.14]{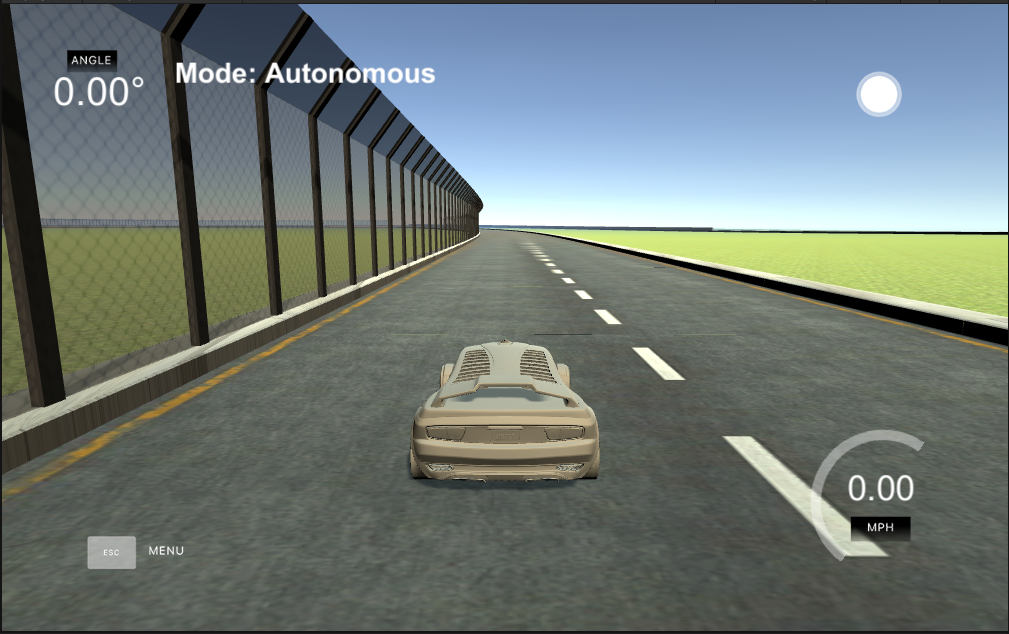}
\caption{Track 2: Simulator and car}
\label{fig:minipage4}
\end{minipage}
\end{figure}

The car has three cameras placed on/near the dashboard, looking left, right, and center/front. In training mode, the simulator can record images from all three cameras, steering, throttle, and speed data at a frequency of 10 Hz.

\subsection{Learning Methodology}
Direct Policy Learning \cite{ross2011reduction} \cite{ross2010efficient}, a type of Imitation Learning is used to train all the Neural Network models in this work. Direct policy learning is a combination of Behavioural Cloning and Data Aggregation. Behavioral cloning is supervised learning with labels provided by the expert on the problem. Data Aggregation \cite{ross2011reduction} is done by having an expert driver in the feedback loop to improve the model's performance iteratively. Figure 5 illustrates the training methodology with an expert in the loop. The expert helps to remove the undesirable driving patterns by correcting/modifying old training data and also provides new training data to improve driving patterns based on significant observations during testing.

\begin{figure}[h]
    \centering
    \includegraphics[scale=0.6]{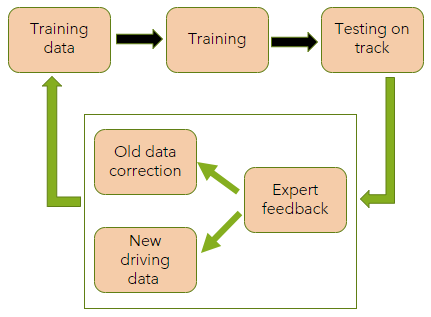}
    \caption{Direct Policy Learning}
    \label{fig:my_label}
\end{figure}


\section{Data Collection Guidelines: A Maximum Speed Perspective}

The key element to racing is driving at high speeds on an optimal path around the racing track. From our experiments, we observe that merely providing driving data on the optimal path is not sufficient for good autonomous driving performance, as the car would typically crash when slightly deviating from the training path. Hence, we break the racing data collection problem down into two stages. First, to learn the ability to drive at high speeds in different situations on track, and then second to learn to drive on an optimal path. We address the first problem in this work, and the second problem of learning the ability to drive on an optimal path will be explored in future work.

\subsection{Driving Data Collection Strategy}

The car on a racing track can run into a wide range of situations; therefore, for it to generalize well on the track, along with amount of driving data, we observe that \textbf{diversity in data is key to learning}. More specifically, we followed the guidelines below to diverse training data.
\begin{itemize}
    \item Driving on both right and left lanes.
    \item Changing lanes at different points of the track.
    \item Driving closer to the edges and coming back towards the center of the track. We observed that this is especially needed for tracks with sharp turns.
\end{itemize}

Table 1 lists the number of laps collected on each track at different constant speeds. We note that the number of laps provided here can be varied with roughly $\pm3$ laps to produce similar results.

\begin{table}[h]
\centering
\caption{Data in Number of laps}
\begin{tabular}{|l|ccc|}
\hline
\diagbox{Track}{Speed} & 80 mph / 60 mph & 50 mph & 30 mph \\
\hline
Track 1   & 65 (80 mph) & 20 & 15\\
Track 2 & 50 (60 mph) & 25 &  15\\
\hline
\end{tabular}
\label{tab:caption}
\end{table}

During the data collection process, the simulator records images with width and height of 320 x 160 from all three cameras, steering values ranging from -1 to +1, throttle values between 0 and 1 and speed data. During training and testing, the input image is cropped and resized to width and height of 160 x 80. We note that the height is cropped to remove the upper part of the image corresponding to the background sky. Data augmentation is used during the training process via introducing vertical and horizontal translation of images and horizontal flipping with inverting the sign of the corresponding steering value.

\subsection{Deep Neural Network}
The deep neural network architecture from Nvidia \cite{bojarski2016end} was used for learning the steering angle. A schematic of the architecture is shown Figure 6.

\begin{figure*}[h]
    \centering
    \includegraphics[width=14cm, height=3.5cm]{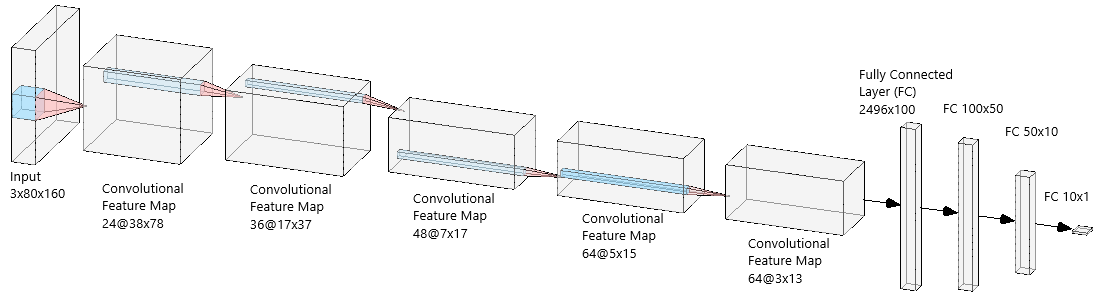}
    \caption{Deep neural network used for learning steering and investigating successful data collection properties. This same network is used for steering learning network when jointly learning steering and throttle.}
    \label{fig:my_label}
\end{figure*}

\subsection{Training and Testing}
Racing is about optimizing performance on a given specific track \cite{milliken1995race}. Our goal is to learn optimal control on a given specific track using a neural network model that generalizes for different and challenging situations during high-speed driving. To achieve that, we are training the model with large and diverse data while testing on the same track. Generalizability of high speed driving skills for different tracks will be explored in future work.

Training in this problem is an iterative process as described above in Section II-B. We first collect data corresponding to a pre-specified number of laps, then train the model using the Mean Squared Error (MSE) loss and the Adaptive Moment estimation (Adam) optimizer \cite{kingma2014adam} with a learning rate of 0.0001. We then evaluate the model's performance on the same track, and test specifically for driving stability. If unstable driving is observed (e.g., the car crashes), we collect more training laps or correct the old data by removing undesired driving patterns and retrain the network. This process continues until the desired maximum speed is maintained in a stable manner. The appropriate number of laps to use for data collection, shown in Table I, is found using this iterative process. The amount of data/number of laps required for learning depends on the track and car speed at which training data is collected. Figure 6 shows the Deep Neural Network architecture used for learning steering control. \textbf{We observe that as the number of laps used for training increases, the number of training epochs has to be increased}. For a number of laps less than 20, the model was trained on 1000 epochs. For a number of laps greater than 20, the model was trained for 2000-3000 epochs. A batch size of 100 was used. The number of epochs during training was found empirically and typically depends on the quality of driving data. We also observe that an important factor which affects the quality of collected data is the availability of smooth turns.

\subsection{Stability and Quality Criteria}

Vision-based driving model's prediction accuracy/error is weakly correlated to the driving performance \cite{codevilla2018offline}. 
Hence, in order to evaluate a model's ability to drive at high speeds in different and challenging situations on a track, we test the model by running for five consecutive laps. During different lap runs, the model has a high chance of running into different and difficult situations that have not been seen before during training. If the model can drive the full five laps with zero collisions then we conclude that it demonstrates generalizability of driving in different situations on a given track.

Having successfully completed five laps, we measure the average lap time over the five laps and further record if the car goes beyond the track edge. These two criteria serve as our measures of driving quality. 

\begin{table}[h]
\begin{center}
    
    \begin{tabular}{| l | l |}
    \hline
    Criterion 1 & Successful completion of 5 laps \\
    \hline
    Criterion 2 & Average Lap Time (ALT) \\
    \hline
    Criterion 3 & Going beyond the track edge line  \\
    \hline
    \end{tabular}
    \caption{Considered Stability and Quality Criteria}
    \label{tab:criteria}
    
\end{center}
\end{table}


\subsection{Results}

Track 1 and Track 2 results are shown in Fig. 7 and 8, respectively.
Empirically, the results demonstrate that the amount of training data governs the maximum speed at which the model will be able to drive in a stable manner.

\begin{figure}[H]
    \centering
    \includegraphics[scale=0.6]{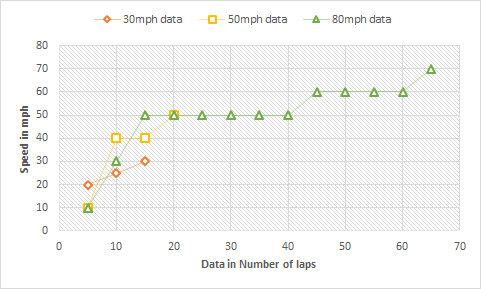}
    \caption{Track 1 Results}
    \label{fig:my_label}
\end{figure}

\begin{figure}[H]
    \centering
    \includegraphics[scale=0.6]{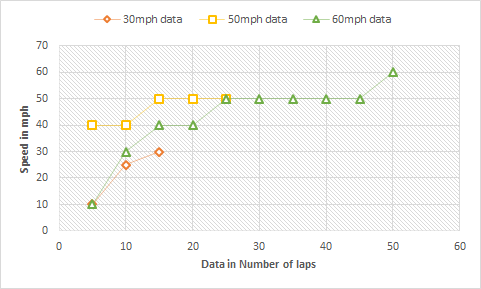}
    \caption{Track 2 Results}
    \label{fig:my_label}
\end{figure}
Insight 1: Establishing \textit{direct relationship between the amount of data and the maximum speed learned by the model}. As we add training data, the model is able to learn higher and higher speeds until it approaches the constant training speed. This relationship helps to address the data-centric question of machine learning in this problem, which is: \textbf{When will adding more data help improve the model's performance, and when should we switch to a model-centric view and explore different architectures to improve performance?}. 

In Tables III and IV, \ding{55} indicates that the model is going beyond the edge line and \ding{51} indicates otherwise. We note that the maximum speed indicated in the tables can vary by $\pm1.5$ mph during a single lap run. The average lap times provided in the tables can also vary by $\pm0.5$ seconds.

Tables III and IV show that as we increase the number of training laps, the maximum stable speed increases and consequently the lap time of the model improves. Further, we observe that the lap time can slightly improve for cases when the added training set does not result in higher maximum speed. The lap time here can be used as one measure of quality of driving when two models are able to successfully complete five laps. Another measure is our \emph{Edge Check}. If the quality of driving is poor, the car oscillates on track touching the edge of track. This can be seen with 15 laps of data on Track 1, where it is able to achieve 50 mph speed but has poor driving quality. The driving quality is improved through the iterative Learning methodology described in Section II-B. We note that increasing the size of the training set by training for more laps does not always lead to improvement in performance, as the quality of training data added has a significant impact. Due to undesired driving patterns in newly collected training data, we observed in several occasions that adding more data degrades the performance of the model. Having an expert in the feedback loop is hence needed to remove/correct the undesired driving patterns before retraining the model. 


\begin{table}[h]
\centering
\caption{Track 1, 80 mph training results}
\begin{tabular}{|p{0.9in}|p{0.4in}|p{0.3in}|p{0.4in}|p{0.3in}|}
\hline
\diagbox{Laps}{Metric} & Speed (mph) & 5 laps & ALT (sec) & Edge \\
\hline
10   & 30 & \checkmark & 195.8 & \checkmark \\
\hline
15   & 50 & \checkmark & 117.7 & \xmark \\
\hline
20   & 50 & \checkmark & 116.5 & \checkmark \\
\hline
35   & 50 & \checkmark & 116.1 & \checkmark \\
\hline
45   & 60 & \checkmark & 99.4 & \xmark \\
\hline
55   & 60 & \checkmark & 98.37 & \checkmark \\
\hline
65   & 70 & \checkmark & 84.19 & \checkmark \\
\hline
\end{tabular}

\label{tab:caption}
\end{table}

\begin{table}[h]
\centering
\caption{Track 2, 60 mph training results}
\begin{tabular}{|p{0.9in}|p{0.4in}|p{0.3in}|p{0.4in}|p{0.3in}|}
\hline
\diagbox{Laps}{Metric} & Speed (mph) & 5 laps & ALT (sec) & Edge \\
\hline
5   & 10 & \checkmark & 839.5 & \checkmark \\
\hline
10   & 30 & \checkmark & 278.2 & \checkmark \\
\hline
15   & 40 & \checkmark & 211.4 & \xmark \\
\hline
20   & 40 & \checkmark & 209.9 & \checkmark \\
\hline
25   & 50 & \checkmark & 158.1 & \xmark \\
\hline
40   & 50 & \checkmark & 158.8 & \checkmark \\
\hline
50   & 60 & \checkmark & 142.4 & \checkmark \\
\hline
\end{tabular}

\label{tab:caption}
\end{table}

Insight 2: \textbf{Training data with higher maximum speed enables stable driving at lower speeds with less data}. 

We note that the frequency of data collection is fixed at 10 Hz, so if 10 laps of data are collected at 30 mph, 50 mph, and 80 mph, then the number of data points collected at these speeds are different as more images will be available for the lower speeds. For example, the amount of data corresponding to 10 laps at 50 mph is the same as that corresponding to 16 laps at 80 mph. 
In Fig. 7 and 8, we observe a dominant pattern - that does not hold only in few exceptions - of being able to reach the lower speeds with less amount of data when training at higher speeds. This pattern can be extreme in some cases. For example, in Fig. 7, we see that 30 mph stable driving is achieved with 10 laps of training data at 80 mph, and the same speed can only be achieved when training at 30 mph via having 15 laps of training data; noting that the amount of data corresponding to 15 laps at 30 mph is equivalent to that of 40 laps at 80 mph.
We are investigating these observations further with more experiments at various speeds and on different tracks for deeper insights and understanding.


\section{Joint Steering and Throttle Prediction}

The ability to make correct steering and throttle predictions without feedback links decreases the response time of the system, which is crucial for autonomous racing. This architecture is an attempt towards achieving that goal.


Here, \textbf{we demonstrate that throttle can be learned using the convolutional layers from the trained steering model without retraining}. Only the subsequent fully connected layers dedicated to throttle prediction have to be trained. Also unlike the work in \cite{8546189}, we show that the throttle can be learned without any LSTM layers, and unlike the work in \cite{Pan-RSS-18}, without any speed feedback as well as without any other feedback links (e.g., as in \cite{8460487}).



\subsection{Training Methodology}

We divide the training methodology for learning steering and throttle into two parts: 1. Training for steering prediction, and then 2. Training for throttle prediction. An overview of the procedure is shown in Algorithm 1.

\begin{figure}[h]
    \centering
    \includegraphics[scale=0.35]{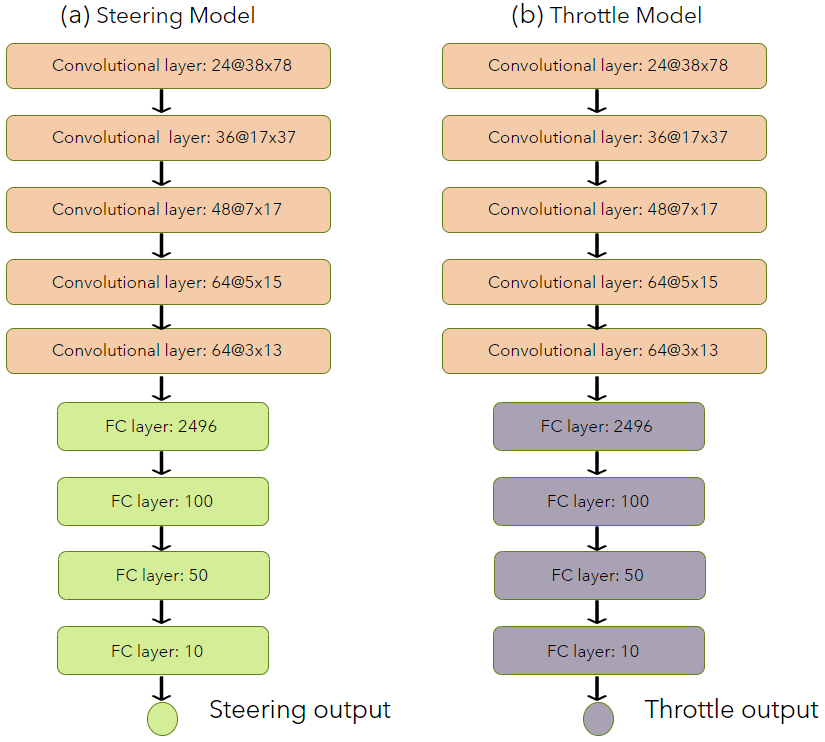}
    \caption{Two separate models for training steering and throttle}
    \label{fig:my_label}
\end{figure}

\subsubsection{Training for Steering Prediction}
The model shown in Fig. 9(a) is used to learn steering. This network is the same as the one shown in Fig. 6. We train this model and improve the performance iteratively using the Direct Policy Learning methodology described in Section II-B.

\subsubsection{Training for Throttle Prediction}
For throttle training, the trained convolutional layers from the steering model are used. We freeze the learning of these convolutional layers, which indicates that we are using the same convolutional features as in the steering model to learn throttle. Different fully connected layers are used for learning throttle and are the only layers which are trained during throttle learning. Fig. 9(b) represents the model used for throttle learning.

\begin{algorithm}[h]
\SetAlgoLined

\textbf{Result: Steering and Throttle Predictive Model}

\While{Criteria 1, 2, and 3 are not satisfied}{

    Get training data OR add more data
    
    SteeringModel $\leftarrow$ NewModel()
    
    \For{\texttt{epoch $\leq$ TotalEpochs}}{
    trainSteering()
    }
    saveTrainedModel()

    ThrottleModel $\leftarrow$ NewModel()
    
    \textbf{ThrottleConvLayers $\leftarrow$ SteeringConvLayers}
    
    \For{\texttt{epoch $\leq$ TotalEpochs}}{
    trainThrottle()
    }
    
    saveTrainedModel()
    
    MergeBothModels()
    
    EvaluatePerformanceOnTrack()
}

\caption{Training procedure for learning steering and throttle}
\label{alg3:train}
\end{algorithm}

\subsubsection{Equivalent Full Model}

We note that when testing the model's performance, conceptually, the equivalent final architecture consists of a combination of both networks shown in Fig. 9(a) and 9(b) into one as shown in Fig. 10. Since the convolutional layer weights are the same, these convolutional layers become the common backbone of the full network. However, we note that training this architecture takes place in two stages as outlined above.

\begin{figure}[h]
    \centering
    \includegraphics[scale=0.4]{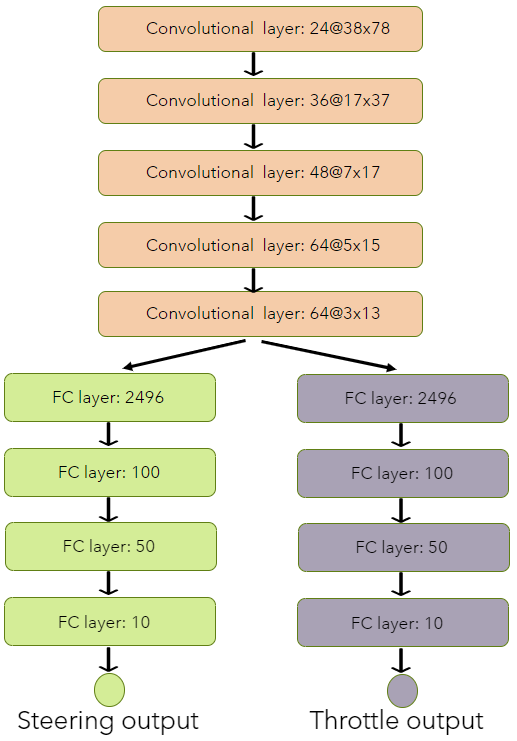}
    \caption{Merged Deep Neural Network architecture for joint prediction of steering and throttle}
    \label{fig:my_label}
\end{figure}



\subsection{Results}
We show results for Track 2, which has sharp turns, to demonstrate the model's ability to control steering and throttle under challenging situations. An additional Lake Track, which is the default track provided in the Udacity simulator, is also used to demonstrate that this neural network architecture and training methodology can be successfully applied to different tracks. At the time of writing this draft, the training and testing process was in progress for the IMS-like Track 1, and the results will be included in future work.

\subsubsection{Track 2}
We provide a video demonstration on Track 2\footnote{Available at https://youtu.be/On0RhWkMLW4}. The video shows the ability of the considered deep neural network shown in Fig. 10 to successfully control steering and throttle. In particular, the model learns to reduce throttle at turns to achieve needed stability and reaches full throttle in straight stretches to achieve a maximum speed of 90 mph.

\subsubsection{Lake Track}
We also provide a video demonstration on the Lake Track\footnote{Available at https://youtu.be/ChaoakkGMgs}.

\section{Conclusion}

In this work, we explored significant challenges related to applying deep-learning-based end-to-end learning solutions for autonomous racing. First, with a data-centric perspective, we explored properties of successful data collection strategies with an emphasis on the maximum reachable speed via stable driving. We demonstrated the details of the data collection process while highlighting the importance of diversity and the amount of data. Important insights were drawn, specifically capturing the relationship between the amount of driving data, in terms of number of training laps, and the model's performance in terms of the maximum achievable speed via stable driving. We used two different tracks for that first study, one similar to the Indy Motor Speedway (IMS), and another with sharp turns. Our second contribution outlined how to jointly learn steering and throttle without feedback links or recurrence (LSTM) in the deep neural network architecture. We demonstrated the success of the proposed deep learning algorithm for joint steering and throttle prediction through the track with sharp turns as well as the Udacity simulator's default Lake Track. The relationship between maximum training speed and the model's steering and throttle prediction performance will be further analyzed in future work.





\section{Acknowledgment}

We would like to extend appreciation to Aref F. Malek, Alec F. Pannunzio, Mikail H. Khan, Abhimanyu Agarwal and Tommy Wygal at Purdue University for their contribution towards building the custom simulator and tracks. We  would also like to thank all members of the Purdue-USMA Indy Autonomous Challenge (IAC) Black \& Gold Autonomous Racing team and the Autonomous Motorsports Purdue (AMP) student club for providing the environment and support that enabled this work, with special thanks to Michael Saxon at the Storm King Group, J. Eric Dietz and Joseph Pekny at Purdue University, Christopher Korpela at the USMA, and Robert Megennis at Andretti Autosport.


\bibliographystyle{plain}
\bibliography{bibliography}

\begin{thebibliography}{10}

\bibitem{bojarski2016end}
Mariusz Bojarski, Davide Del~Testa, Daniel Dworakowski, Bernhard Firner, Beat
  Flepp, Prasoon Goyal, Lawrence~D Jackel, Mathew Monfort, Urs Muller, Jiakai
  Zhang, et~al.
\newblock End to end learning for self-driving cars.
\newblock {\em arXiv preprint arXiv:1604.07316}, 2016.

\bibitem{codevilla2018offline}
Felipe Codevilla, Antonio~M Lopez, Vladlen Koltun, and Alexey Dosovitskiy.
\newblock On offline evaluation of vision-based driving models.
\newblock In {\em Proceedings of the European Conference on Computer Vision
  (ECCV)}, pages 236--251, 2018.

\bibitem{8460487}
Felipe Codevilla, Matthias Müller, Antonio López, Vladlen Koltun, and Alexey
  Dosovitskiy.
\newblock End-to-end driving via conditional imitation learning.
\newblock In {\em 2018 IEEE International Conference on Robotics and Automation
  (ICRA)}, pages 4693--4700, 2018.

\bibitem{he2016deep}
Kaiming He, Xiangyu Zhang, Shaoqing Ren, and Jian Sun.
\newblock Deep residual learning for image recognition.
\newblock In {\em Proceedings of the IEEE conference on computer vision and
  pattern recognition}, pages 770--778, 2016.

\bibitem{hecker2018end}
Simon Hecker, Dengxin Dai, and Luc Van~Gool.
\newblock End-to-end learning of driving models with surround-view cameras and
  route planners.
\newblock In {\em Proceedings of the european conference on computer vision
  (eccv)}, pages 435--453, 2018.

\bibitem{juliani2018unity}
Arthur Juliani, Vincent-Pierre Berges, Ervin Teng, Andrew Cohen, Jonathan
  Harper, Chris Elion, Chris Goy, Yuan Gao, Hunter Henry, Marwan Mattar, et~al.
\newblock Unity: A general platform for intelligent agents.
\newblock {\em arXiv preprint arXiv:1809.02627}, 2018.

\bibitem{kingma2014adam}
Diederik~P Kingma and Jimmy Ba.
\newblock Adam: A method for stochastic optimization.
\newblock {\em arXiv preprint arXiv:1412.6980}, 2014.

\bibitem{krizhevsky2012imagenet}
Alex Krizhevsky, Ilya Sutskever, and Geoffrey~E Hinton.
\newblock Imagenet classification with deep convolutional neural networks.
\newblock {\em Advances in neural information processing systems},
  25:1097--1105, 2012.

\bibitem{lin2014microsoft}
Tsung-Yi Lin, Michael Maire, Serge Belongie, James Hays, Pietro Perona, Deva
  Ramanan, Piotr Doll{\'a}r, and C~Lawrence Zitnick.
\newblock Microsoft coco: Common objects in context.
\newblock In {\em European conference on computer vision}, pages 740--755.
  Springer, 2014.

\bibitem{milliken1995race}
William~F Milliken, Douglas~L Milliken, et~al.
\newblock {\em Race car vehicle dynamics}, volume 400.
\newblock Society of Automotive Engineers Warrendale, PA, 1995.

\bibitem{Pan-RSS-18}
Yunpeng Pan, Ching-An Cheng, Kamil Saigol, Keuntaek Lee, Xinyan Yan, Evangelos
  Theodorou, and Byron Boots.
\newblock Agile autonomous driving using end-to-end deep imitation learning.
\newblock In {\em Proceedings of Robotics: Science and Systems}, Pittsburgh,
  Pennsylvania, June 2018.

\bibitem{ross2010efficient}
St{\'e}phane Ross and Drew Bagnell.
\newblock Efficient reductions for imitation learning.
\newblock In {\em Proceedings of the thirteenth international conference on
  artificial intelligence and statistics}, pages 661--668. JMLR Workshop and
  Conference Proceedings, 2010.

\bibitem{ross2011reduction}
St{\'e}phane Ross, Geoffrey Gordon, and Drew Bagnell.
\newblock A reduction of imitation learning and structured prediction to
  no-regret online learning.
\newblock In {\em Proceedings of the fourteenth international conference on
  artificial intelligence and statistics}, pages 627--635. JMLR Workshop and
  Conference Proceedings, 2011.

\bibitem{ILSVRC15}
Olga Russakovsky, Jia Deng, Hao Su, Jonathan Krause, Sanjeev Satheesh, Sean Ma,
  Zhiheng Huang, Andrej Karpathy, Aditya Khosla, Michael Bernstein,
  Alexander~C. Berg, and Li~Fei-Fei.
\newblock {ImageNet Large Scale Visual Recognition Challenge}.
\newblock {\em International Journal of Computer Vision (IJCV)},
  115(3):211--252, 2015.

\bibitem{szegedy2015going}
Christian Szegedy, Wei Liu, Yangqing Jia, Pierre Sermanet, Scott Reed, Dragomir
  Anguelov, Dumitru Erhan, Vincent Vanhoucke, and Andrew Rabinovich.
\newblock Going deeper with convolutions.
\newblock In {\em Proceedings of the IEEE conference on computer vision and
  pattern recognition}, pages 1--9, 2015.

\bibitem{8099859}
Huazhe Xu, Yang Gao, Fisher Yu, and Trevor Darrell.
\newblock End-to-end learning of driving models from large-scale video
  datasets.
\newblock In {\em 2017 IEEE Conference on Computer Vision and Pattern
  Recognition (CVPR)}, pages 3530--3538, 2017.

\bibitem{8546189}
Zhengyuan Yang, Yixuan Zhang, Jerry Yu, Junjie Cai, and Jiebo Luo.
\newblock End-to-end multi-modal multi-task vehicle control for self-driving
  cars with visual perceptions.
\newblock In {\em 2018 24th International Conference on Pattern Recognition
  (ICPR)}, pages 2289--2294, 2018.

\end{thebibliography}

\end{document}